\documentclass{article}

\usepackage[preprint]{corl_2026} 

\usepackage{booktabs}
\usepackage{multirow}
\usepackage{amsmath}
\usepackage{amssymb}
\usepackage{graphicx}
\usepackage{xcolor}
\usepackage{subcaption}
\usepackage{wrapfig}
\usepackage{enumitem}
\usepackage{float}
\usepackage{microtype}
\usepackage{placeins}
\usepackage{tikz}
\usepackage{bbm}
\usetikzlibrary{arrows.meta, positioning, calc, fit, backgrounds}
\usepackage[capitalize,nameinlink]{cleveref}

\newcommand{\method}{RoboNaldo}
\newcommand{\cmark}{\checkmark}
\newcommand{\xmark}{$\times$}

\title{\method: Accurate, Stable and Powerful Humanoid Soccer Shooting via Motion-Guided Curriculum Reinforcement Learning}

\author{
\normalfont
  \textbf{Yichao Zhong}$^{1,*}$ \quad
  \textbf{Yidan Lu}$^{1,*}$ \quad
  \textbf{Yuhang Lu}$^{1}$ \quad
  \textbf{Tianyang Tang}$^{1}$\\ 
  \textbf{Haoguang Mai}$^{1}$ \quad
  \textbf{Yixuan Pan}$^{3}$ \quad
  \textbf{Tianyu Li}$^{3,\dagger}$ \quad
  \textbf{Li Chen}$^{1,\dagger}$ \\
  \textbf{Jingbo Wang}$^{2}$ \quad
  \textbf{Zhongyu Li}$^{2,\ddagger}$ \quad
  \textbf{Peng Lu}$^{1,\ddagger}$ \quad
  \textbf{Hongyang Li}$^{3,1,\ddagger}$ \\[2mm]
  {\normalfont $^{1}$The University of Hong Kong \quad
  $^{2}$The Chinese University of Hong Kong} \quad
  {\normalfont $^{3}$Archon Robotics} \\
  {\normalfont $^{*}$Equal contribution \quad
  $^{\dagger}$Project co-lead \quad
  $^{\ddagger}$Equal advising} \\[2mm]
  {\normalfont\ttfamily yichao.zhong@connect.hku.hk, hongyang@hku.hk}
}

\begin{document}
\maketitle


{\centering
  \includegraphics[width=0.95\linewidth,trim=0 0 0 75,clip]{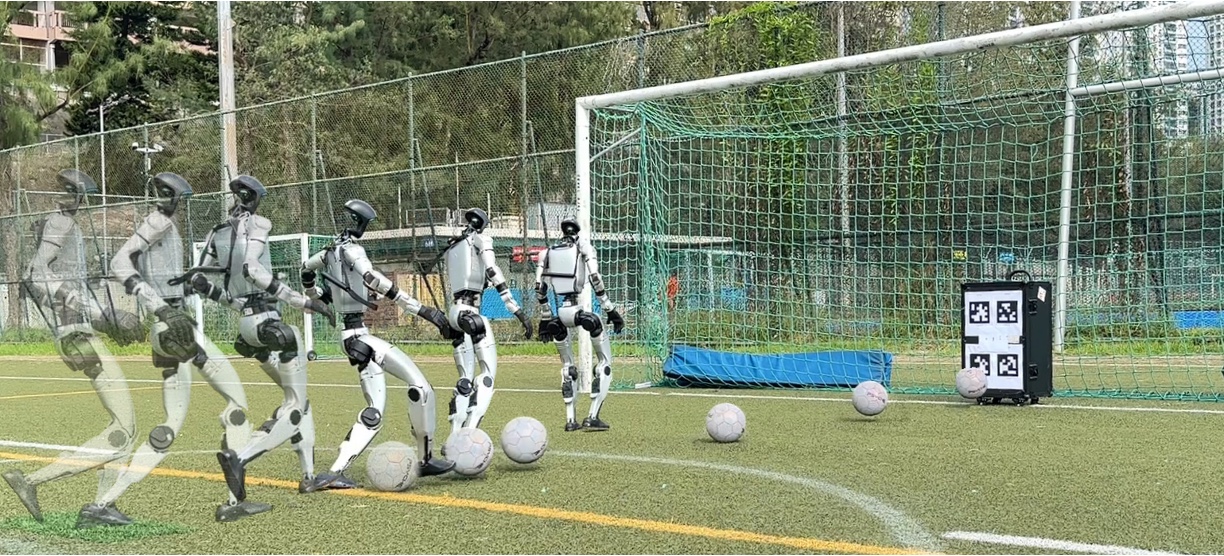}\par
  \captionof{figure}{
    \textbf{\method{}: accurate, stable and powerful whole-body soccer shooting on a real grassy soccer field.}
    Ghost-overlaid time sequence of a Unitree G1 humanoid executing a directed penalty kick. 
  }
  \label{fig:teaser}
}

\begin{abstract}
Elite humanoid soccer shooting requires whole-body stability, high-impulse whole-body interactions, and accuracy to targets.
Motion tracking-driven reinforcement learning (RL) provides stability in whole-body movement coordination, but a fixed reference makes it hard to adapt to varied ball positions and strike timings; in contrast, task reward-driven RL struggles to explore and discover valid kicks from scratch. We therefore introduce \textbf{\method{}}, a three-stage motion-guided curriculum RL framework for high-impulse humanoid interaction. A single human-kick reference is used as a scaffold and progressively shifts optimization towards shooting performance.
The curriculum first learns stable whole-body kicking prior, then adapts the kick to free-kick settings where the ball is stationary at random positions, and finally extends it to moving-ball shooting through a locomotion-command and kick-trigger interface. A high-level heuristic planner controls this interface during training, while alternative high-level controllers can drive the same low-level policy at inference.
In simulation, \method{} demonstrates free-kick shot error 48.6\% lower and shoot velocity $2.96\times$ than prior work baselines. In real world on a Unitree G1 with onboard perception, \method{} attains 0.73\,m and 0.86\,m average target shooting error from 3\,m away in free-kick and moving-ball cases, accordingly. And the post-contact ball velocity reaches 13.10\,m/s, which is 59--71\% of reported professional open-play shot speed.
Project page: \href{https://opendrivelab.com/RoboNaldo}{\textcolor{magenta}{https://opendrivelab.com/RoboNaldo}}.
\end{abstract}



\section{Introduction}
\label{sec:intro}

Soccer shooting provides a compact but demanding benchmark for athletic humanoid interaction. A successful shot requires the robot to direct the ball towards a target while generating a high-velocity strike and maintaining whole-body stability. This couples several challenging control problems, including single-leg balance, millisecond-scale foot–ball contact, and generalization across diverse ball and target configurations. Unlike sustained-contact tasks such as lifting or pushing, shooting relies on a brief, high-impulse collision generated by an end-effector that simultaneously contributes to whole-body support and balance. 

Recent humanoid soccer systems~\citep{haarnoja2024learning,wang2025visiondriven,wang2026humanx,kong2026paid} have made rapid progress in locomotion, ball chasing, and goal-directed kicking. However, their shooting evaluations remain limited in accuracy, power, and deployment scope. For example, the ball direction cosine metric in~\citep{kong2026paid} exceeds 90\% even when the shot-on-goal error is 2.4\,m from a 5\,m shooting distance. Among prior systems, only~\citep{wang2026humanx} considers moving-ball kicks, but it does not evaluate shooting accuracy or kick power, and the system is not demonstrated in outdoor deployment. These limitations point to a broader methodological challenge: existing ingredients for humanoid kicking provide either motion-level stability or task-level objectives, but they do not directly resolve the contact selection and timing decisions required for accurate shooting. Motion priors make the robot stable enough to kick, but a fixed reference cannot choose a new contact point, aim direction, or strike time~\citep{peng2018deepmimic,kong2026paid,wang2026humanx}.
Task rewards specify where the ball should go, but pure RL must discover balance, swing coordination, contact, and aim from sparse delayed feedback; AMP controllers~\citep{peng2021amp,wang2025visiondriven} encourage human-like behavior but still leave when-to-kick and where-to-aim hard to supervise.
As a result, prior work has not jointly demonstrated \emph{stability}, \emph{accuracy}, \emph{power}, and \emph{generalizability} for humanoid soccer shooting.

We present \textbf{\method}, a learning framework for accurate, stable and high-impulse humanoid interactions like soccer shooting.
The key idea is to stage the problem according to what each learning signal can reliably provide. In \emph{Stage~1}, 
motion tracking learns a stable kicking prior that captures balance and swing structure. In \emph{Stage~2}, shooting rewards and diverse free-kick configurations teach the policy to adapt its strike for target-directed accuracy. In \emph{Stage~3}, locomotion-command/kick-trigger interface converts moving-ball shooting into approach control plus contact-timing decisions. After training, the Stage~3 policy can be driven by a heuristic controller or by other high-level commands, such as human commands or a co-trained high-level neural policy.

We validate \method{} in both simulation and real-world experiments.
In simulation, Stage~2 free-kicks achieve 0.899\,m average error from 5\,m, 65.5\% of the shot within 1\,m error, and 14.79\,m/s ball speed, giving 0.5$\times$ the error and 2.96$\times$ the speed of previous work; Stage~3 preserves high-power moving-ball shooting with 63.3\% shots within 1\,m error.
On hardware, \method{} runs fully onboard on a Unitree G1 with LiDAR-camera perception and performs directed shots on a real grassy soccer field, as shown in~\cref{fig:teaser}. It achieves 0.73\,m average low-target free-kick error from 3\,m, ball speeds up to 13.10\,m/s, and 74\% contact in human-passed moving-ball trials.
These results show that the proposed curriculum learns a shooting policy that is stable, accurate, and powerful across the evaluated free-kick and moving-ball regimes. Our contributions can be summarized as follows:
\begin{itemize}
  \item We demonstrate powerful, and accurate real-world humanoid soccer shots with generalization and stability across ball velocities and positions.
  \item We propose a staged curriculum for high-impulse humanoid interactions. Starting from learning a motion prior, then learns contact and timing adaptations, converting general moving ball shooting into a learnable problem. 
  \item \method{} achieves accurate and powerful shooting, attaining 0.73\,m and 0.86\,m mean shot error in free-kick and moving-ball cases; the ball speeds up to 13.10\,m/s (47.2\,km/h), reaching ${\sim}$71\% of female and ${\sim}$59\% of professional male open-play shot speed at the UEFA European Championships~\citep{delacruz2026shotvelocity}.
\end{itemize}


\section{Related Work}
\label{sec:related}

\paragraph{Humanoid soccer and the shooting gap.}
RL has produced sports robots for quadruped soccer~\citep{su2025quadruped}, table tennis~\citep{su2025hitter}, badminton~\citep{ma2025badminton}, and egocentric humanoid kicking and goalkeeping~\citep{haarnoja2024learning,li2025kicktraj,wang2026humanx,wang2025visiondriven,xu2026striker,kong2026paid,ren2025goalkeeper}.
A useful shot demands stability, point-level accuracy, high post-contact speed, and generalization across ball configurations.
Tracking controllers inherit stable coordination but constrain strike timing; AMP relaxes hard reference constraints but weakly supervises aim; pure task RL struggles with sparse rewards in a large humanoid control space, leading to credit-assignment issues which drops learning performances.
The closest concurrent system, PAiD~\citep{kong2026paid}, shares staged training and egocentric sensing but targets goal-region entry, not point-level placement; its cosine metric cannot resolve sub-metre accuracy differences.
\method{} targets point-level accuracy, high speed, moving-ball shooting, onboard sensing, and outdoor deployment jointly (\cref{tab:related}).

\begin{wraptable}{r}{0.5\linewidth}
  \vspace{-\intextsep}
  \centering
  \caption{\textbf{Comparison of different humanoid soccer
systems.} We compare target-point accuracy (Point-acc), reported ball speed (Speed), moving-ball shooting (Moving), egocentric sensing (Ego), and outdoor demonstration (Out). $\dagger$ denotes concurrent work.}
  \label{tab:related}
  \vspace{-0.1em}
  {\scriptsize
  \setlength{\tabcolsep}{2pt}
  \renewcommand{\arraystretch}{0.95}
  \begin{tabular*}{\linewidth}{@{\extracolsep{\fill}}lccccc@{}}
  \toprule
  \multirow{2}{*}{Method}
    & \multicolumn{3}{c}{Shooting capability}
    & \multicolumn{2}{c}{Deployment} \\
  \cmidrule(lr){2-4}\cmidrule(lr){5-6}
    & Point-acc & Speed & Moving & Ego & Out \\
  \midrule
  STOFT~\citep{li2025kicktraj}                    & \xmark & \xmark & \xmark & \xmark & \xmark \\
  Reactive~\citep{wang2025visiondriven} & \xmark & \xmark & \xmark & \cmark & \cmark \\
  Striker~\citep{xu2026striker}         & \xmark & \xmark & \xmark & \cmark & \xmark \\
  HumanX$^\dagger$~\citep{wang2026humanx}                   & \xmark & \xmark & \cmark & \xmark & \xmark \\
  PAiD$^\dagger$~\citep{kong2026paid}             & \xmark & \xmark & \cmark & \cmark & \cmark \\
  \textbf{\method{}(Ours)}                              & \cmark & \cmark & \cmark & \cmark & \cmark \\
  \bottomrule
\end{tabular*}
  }
  \vspace{-\intextsep}
  \vspace{-4mm}
\end{wraptable}

\paragraph{Humanoid whole-body object interaction.}
Contact-rich loco-manipulation systems~\citep{weng2025hdmi,zhao2025resmimic,kalaria2025dreamcontrol,zhang2026falcon,xu2025intermimic} assume dwell-time contact and accumulate interaction rewards across multiple steps.
Soccer shooting instead has a sub-10\,ms impulse and delayed ball--target feedback; \method{} addresses this with an Instant Interaction Reward and a Densified Shooting Reward that extrapolates post-contact ball state.

\paragraph{Curriculum reinforcement learning.}
Curriculum RL ramps task difficulty~\citep{bengio2009curriculum,narvekar2020curriculum} and is common in legged locomotion~\citep{rudin2022learning}.
For soccer shooting, increasing ball-position diversity before the policy can kick yields no reward signal; \method{} co-designs reward weights and task difficulty across three behavior-triggered stages.


\section{Method}
\label{sec:method}

\begin{figure}[!h]
  \centering
  \vspace{-25mm}
  \includegraphics[width=1\linewidth]{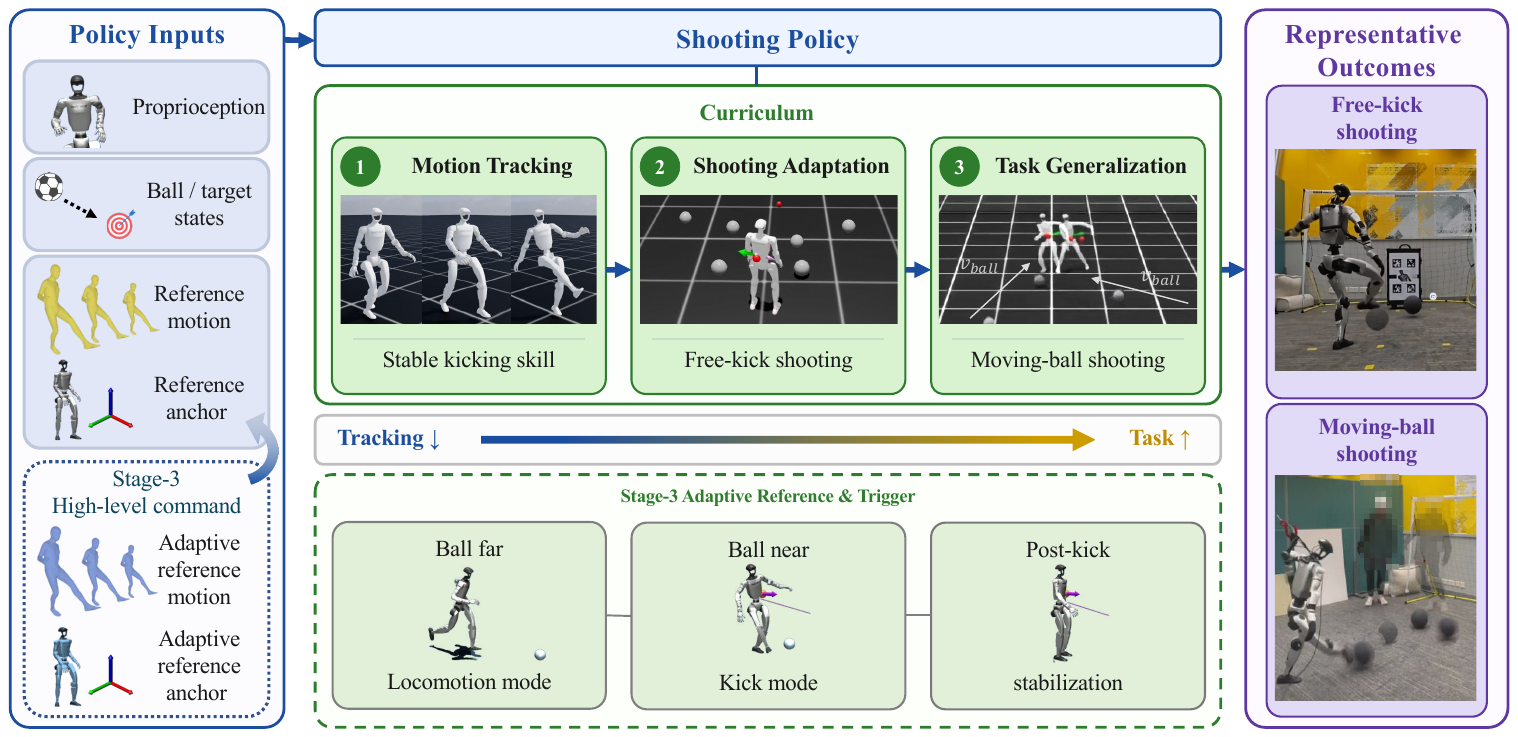}
  \vspace{-25mm}
  \caption{\textbf{\method{} motion-guided curriculum.} A single low-level policy is trained sequentially across three stages.  Stage~1 learns a stable kick by tracking a retargeted human motion.  Stage~2 introduces shooting task rewards to adapt the kick for accurate stationary-ball free-kicks.  Stage~3 replaces the reference motion and anchor cue with high-level locomotion and kick-trigger decision commands, enabling the policy to approach, trigger, strike, and stabilize when shooting moving balls.}
  \label{fig:pipeline}
  \vspace{-3mm}
\end{figure}

\method{} trains a unified policy for locomotion and soccer shooting using a \textbf{three-stage task curriculum} (\cref{fig:pipeline}). We begin by formulating the task and specifying the observation space (Sec.\cref{sec:task}). We then introduce the three-stage curriculum that progressively develops shooting skills (\cref{sec:curriculum}). Finally, we present the reward design that encourages accurate, powerful, and stable shots (~\cref{sec:reward}).

\subsection{Task Formulation}
\label{sec:task}
\paragraph{Soccer shooting task.}
The task is specified by the ball state and a static target point on the goal line.  We express ball and target observations in the robot frame for policy input, and evaluate shot accuracy in the world frame. Free-kick shooting refers to the stationary-ball setting, where the ball starts from rest at a sampled location and the robot must strike it toward the target. Moving-ball shooting extends this setting by assigning the ball an incoming velocity and randomized arrival timing; the robot must therefore both reach a feasible contact configuration and trigger the kick at the appropriate time.
Let $p_b(t)$ denote the ball position after contact and let $p_t$ denote the target point.  A trial is successful if
$\min_{t \geq t_c} \|p_b(t)-p_t\|_2 \leq d_{\rm thresh},$
where $t_c$ is the first valid foot--ball contact.  This post-contact definition
ensures that accuracy is measured only after the robot has launched or redirected
the ball.
\paragraph{Policy observation.}
The policy receives motion-reference cues, proprioceptive history, previous actions, and ball/target observations.  In Stages~1 and 2, the reference-anchor cue specifies the motion-reference anchor used for tracking.  In Stage~3, this anchor cue is replaced by the locomotion command used by the moving-ball planner; the kick-trigger signal determines when the reference switches from locomotion to kicking. The full per-dimension observation vector is given in~\cref{app:obs}.

\subsection{Curriculum Design}
\label{sec:curriculum}


\method{} uses a three-stage curriculum to separate the main difficulties of humanoid shooting: acquiring a stable whole-body kick, adapting the strike to different stationary-ball configurations, and timing the kick for moving balls. The stages are trained sequentially, with each stage initialized from the previous checkpoint.

\textbf{Stage~1: Shooting Motion Tracking.}
The reference is a side-foot kick retargeted from human video via GVHMR~\citep{shen2024gvhmr} and GMR~\citep{araujo2025gmr}.
Following BeyondMimic~\citep{liao2025beyondmimic}, the policy imitates this reference without ball or task rewards, establishing coordination before aim/contact adaptation.
Stages advance after task reward plateaus, action-noise standard deviation converges, and behavior emerges.

\textbf{Stage~2: Shooting Adaptation.}
Stage~2 introduces the ball, the target, and the shooting rewards while randomizing the ball spawn position around the nominal contact location.  The policy must therefore adapt the demonstrated kick rather than simply replay it: the approach, contact point, strike direction, and impact speed all depend on the sampled ball and target.  The resulting checkpoint serves as the free-kick policy and provides the initialization for moving-ball training.

\textbf{Stage~3: Task Generalization.}
Moving-ball shooting adds a temporal-alignment challenge.  At realistic pass speeds, useful foot--ball contact occurs within only a few control steps, while the motion reference still imposes a nominal swing phase.  Directly continuing Stage~2 under fast incoming-ball randomization provides little reliable contact signal.  Stage~3 therefore makes timing explicit by introducing a high-level locomotion command and a kick trigger, dividing each episode into locomotion, kick, and post-kick stabilization modes.

\textbf{Heuristic planner.}
Before contact, the policy locomotes toward the ball using discounted motion-tracking rewards, goal-directed velocity, and gait-quality terms. The planner steers the anchor toward the predicted ball position $\tilde{\mathbf{p}}b{=}\mathbf{p}{b,\perp}{+}\mathbf{v}{b,\perp}T_h$ while aligning the robot with the ball direction. It triggers the kick when the estimated closest body-frame approach distance $d{\min}$ to the nominal foot-strike offset satisfies
$(d_{\min} < r_{\rm thr}) \wedge (z_b < h_{\max}) \wedge (\phi < \phi_{\rm kick})$,
meaning that the ball is reachable, kickable, and the kick phase has not started. The motion clock is then snapped to $\phi_{\rm kick}$, with small training-time phase jitter for robustness, and the policy switches to full-weight kick tracking. After contact, a post-kick stabilization reference is held for $T_{\rm stab}$ steps before resetting the clock, discouraging unstable impacts that score high shot rewards but fail immediately after the kick.



\paragraph{High-level control interface.}
The locomotion command and kick-trigger signal form a modular interface between the high-level planner and the low-level shooting policy.  The rule-based planner used during training can therefore be replaced at deployment by another high-level controller without retraining the low-level policy.

\subsection{Reward Design}
\label{sec:reward}

The per-step reward is $r = r_\text{motion}w_\text{motion}+r_\text{task}w_\text{task}+r_\text{reg}w_\text{reg}$, the complete formulas of each reward and stage-wise weights are shown in~\cref{app:reward}.

\noindent
\paragraph{Motion tracking reward.}
$r_\text{motion}$ follows BeyondMimic~\citep{liao2025beyondmimic}: exponential-kernel tracking on 14 body parts, with extra foot-velocity weight for contact sensitivity.
\paragraph{Regularization reward.}
$r_\text{reg}$ penalizes action rate, smoothness, foot slip, airborne time, joint velocities/torques, and joint-limit violations~(\cref{tab:curriculum,app:reward}).
\paragraph{Soccer shooting task reward.}
$r_\text{task}$ covers foot/CoM approach, ball speed, contact orientation, the Instant Interaction Reward (\cref{eq:interact}), and a densified shot-placement reward.
For shot placement, we combine a sparse hit signal
$r_\text{sparse goal}{=}\mathbf{1}[\hat{d}\le d_\text{threshold}]$
with a dense distance signal
$r_\text{dense goal}{=}\exp(-\hat{d}^{2}/\sigma_\text{goal}^{2})$,
where
$\hat{d}{=}\min_{t'\le t}\|\mathbf{p}_\text{ball}-\mathbf{p}_\text{target}\|$,
with $d_\text{threshold}{=}0.5$\,m in Stage~2 and $0.35$\,m in Stage~3.
To provide earlier aiming feedback than the episode-minimum distance alone, we also ballistically extrapolate the post-contact ball state to its goal-line crossing and reward the predicted target error at each post-contact step, thereby densifying the long-horizon placement objective without changing the evaluation metric.
Stage~3 additionally activates velocity-tracking and feet air-time locomotion terms.

\paragraph{Instant interaction reward.}
HDMI~\citep{weng2025hdmi} uses a position--force product, which is effective for sustained contact but collapses for soccer shooting: meaningful foot--ball contact lasts only 3--5 physics steps at 200\,Hz.
\method{} instead rewards the full contact lifecycle--approach, impact, and post-contact ball outcome:
\begin{equation}
  \label{eq:interact}
  r_\text{interact}=\bigl(R_\text{contact}+R_\text{goal}\bigr)\bigl(R_\text{vel}+R_\text{force}\bigr)/4,
\end{equation}
where $R_\text{contact}$ shapes approach, $R_\text{force}$ and $R_\text{vel}$ demand high-impulse impact, and $R_\text{goal}$ preserves result-level aiming gradients.
The additive groups prevent reward collapse around the brief contact window while still penalizing weak, slow, or misdirected strikes.


\section{Simulation Experiments}
\label{sec:sim}

\paragraph{Setup and metrics.}
We train in Isaac Lab~\citep{mittal2023orbit} with 4096 parallel environments on an NVIDIA RTX 4090 and evaluate on 16{,}384 held-out episodes.
Targets are sampled on an $8\,\mathrm{m}\times2\,\mathrm{m}$ goal plane 5\,m ahead. In \emph{free-kick}, a stationary ball is sampled in a $1\,\mathrm{m}\times1\,\mathrm{m}$ frontal square; in \emph{moving-ball shooting}, a ball is launched at $0$--$5\,\mathrm{m/s}$ toward a random point in a $2\,\mathrm{m}\times2\,\mathrm{m}$ frontal square.
Reported metrics are alive rate, shot error (mean per-episode nearest ball-target distance), success rates (error $<$ 0.5/1.0\,m), peak ball speed, and ball-contact rates.

\paragraph{Experimental results.}
We evaluate \method{} against naive and existing methods in free-kick and moving-ball regimes.
Pure PPO/AMP rarely make stable directed contact; pure motion tracking remains a replayed kick.
Stage~2 learns accurate, powerful free-kicks but does not zero-shot to moving balls, while Stage~3 generalizes to both regimes.
Degradation concentrates at extreme lateral and high targets, as visualized in the target-conditioned heatmaps in \cref{fig:results}.

\begin{table}[!t]
  \centering\small
  \caption{\textbf{Simulation results and ablations.}
     Shot err. is nearest ball--target distance; suc. denotes success rates with shot error less than 0.5/1.0 \,m; $v_\mathrm{ball}^{max}$ is peak post-contact ball speed. Free-kick uses a stationary ball; moving-ball shooting redirects a ground ball launched at 0--5 m/s. Contact denotes the ratio of envs where the robot contacts with ball. Ablated variants warm-start from the corresponding \method{} stage checkpoint. Nominal policies and metrics are bold.}
  \resizebox{\linewidth}{!}{%
  \begin{tabular}{llcccccc}
    \toprule
    Entry & Regime & 0.5m suc.  (\%) $\uparrow$ & 1.0m suc. (\%) $\uparrow$ & Shot err. (m) $\downarrow$ & $v^{max}_\text{ball}$ (m/s) $\uparrow$ & Contact (\%) $\uparrow$ & Alive (\%) $\uparrow$ \\
    \midrule
    \multicolumn{8}{l}{\emph{Main comparison: free-kick}} \\
    \quad PPO~\cite{schulman2017proximal}
      & Free-kick & $0.0\pm0.0$  & $0.0\pm0.0$  & $4.721\pm0.007$ & $0.780\pm0.003$ & $0.0\pm0.0$   & $0.0\pm0.0$ \\
    \quad AMP~\cite{peng2021amp}
      & Free-kick & $0.9\pm0.1$  & $3.3\pm0.1$  & $3.733\pm0.004$ & $1.771\pm0.051$ & $38.6\pm2.8$  & $41.6\pm2.3$ \\
    \quad PAiD~\citep{kong2026paid}
      & Free-kick & $8.2\pm0.7$  & $19.5\pm0.5$  & $1.850\pm0.026$ & $4.986\pm0.017$ & $67.7\pm0.3$  & $67.3\pm0.5$ \\
    \quad Stage~1 motion tracking
      & Free-kick & $2.6\pm0.1$  & $9.2\pm0.3$  & $2.835\pm0.008$ & $1.353\pm0.006$ & $27.9\pm0.6$  & $99.9\pm0.1$ \\
    \quad \textbf{\method{} stage~2}
      & Free-kick & $28.8\pm1.3$ & $65.5\pm0.6$ & {\boldmath$0.899\pm0.012$} & {\boldmath$14.792\pm0.011$} & {\boldmath$82.1\pm0.3$} & {\boldmath$100.0\pm0.0$} \\
    \quad \method{} stage~3
      & Free-kick & {\boldmath$36.3\pm0.1$} & {\boldmath$66.5\pm0.2$} & $1.055\pm0.012$ & $14.178\pm0.034$ & $81.1\pm0.1$ & $99.7\pm0.1$ \\
    \addlinespace[3pt]
    \multicolumn{8}{l}{\emph{Main comparison: moving-ball shooting}} \\
    \quad PPO~\cite{schulman2017proximal}
      & Moving & $0.6\pm0.1$  & $2.3\pm0.1$  & $3.962\pm0.004$ & $4.251\pm0.013$ & $0.5\pm0.1$   & $0.0\pm0.0$ \\
    \quad AMP~\cite{peng2021amp}
      & Moving & $0.9\pm0.1$  & $3.5\pm0.1$  & $3.575\pm0.015$ & $4.334\pm0.012$ & $12.5\pm0.2$  & $15.7\pm0.3$ \\
    \quad PAiD~\citep{kong2026paid}
      & Moving & $3.5\pm0.4$  & $9.3\pm0.6$  & $2.149\pm0.034$ & $5.030\pm0.005$ & $50.6\pm0.6$  & $48.1\pm0.5$ \\
    \quad Stage~2 zero-shot
      & Moving & $1.9\pm0.1$  & $7.2\pm0.1$  & $3.159\pm0.003$ & $4.250\pm0.009$ & $15.3\pm0.4$  & $37.4\pm0.2$ \\
    \quad \textbf{\method{} stage~3}
      & Moving & {\boldmath$32.4\pm0.6$} & {\boldmath$63.3\pm0.6$} & {\boldmath$1.131\pm0.004$} &  {\boldmath$13.875\pm0.004$} & {\boldmath$79.2\pm0.1$} & {\boldmath$98.8\pm0.1$} \\
    \addlinespace[3pt]
    \multicolumn{8}{l}{\emph{Ablations: curriculum}} \\
    \quad Stage~0$\!\to\!$2
      & Free-kick & $0.0\pm0.0$ & $0.3\pm0.1$ & $4.302\pm0.016$ & $4.586\pm0.210$ & $18.8\pm0.5$ & $1.2\pm0.1$ \\
    \quad Stage~1$\!\to\!$3
      & Moving & $1.7\pm0.2$ & $6.4\pm0.2$ & $3.453\pm0.033$ & $4.196\pm0.005$ & $13.9\pm0.8$ & $65.3\pm0.5$ \\
    \quad Stage~2$\!\to\!$3 w/o planner
      & Moving & $0.9\pm0.3$ & $3.9\pm0.3$ & $3.283\pm0.024$ & $4.710\pm0.006$ & $19.1\pm1.1$ & $18.9\pm1.0$ \\
    \addlinespace[3pt]
    \multicolumn{8}{l}{\emph{Ablations: mechanism}} \\
    \quad W/o adaptive sampling
      & Moving & $4.5\pm0.2$ & $12.6\pm0.3$ & $3.398\pm0.011$ & {$12.084\pm0.064$} & $28.9\pm0.2$ & $32.8\pm2.0$ \\
    \quad W/o stabilization
      & Moving & $15.3\pm0.2$ & $25.6\pm0.3$ & $3.107\pm0.036$ & $12.967 \pm 0.023$ & $34.1\pm1.0$ & $24.4\pm1.2$ \\
    \addlinespace[3pt]
    \multicolumn{8}{l}{\emph{Ablations: interaction reward}} \\
    \quad HDMI-style reward
      & Free-kick & $2.5\pm0.1$ & $8.9\pm0.2$ & $2.865\pm0.001$ & $2.715\pm0.012$ & $45.0\pm0.4$ & $98.7\pm0.2$ \\
    \bottomrule
  \end{tabular}}
  \vspace{4pt}
  \label{tab:main}
  \label{tab:ablation}
  \vspace{-5mm}
\end{table}

\paragraph{Ablations.}
\label{sec:ablation}
We ablate the staged curriculum, Stage~3 mechanism, and interaction reward in \cref{tab:ablation}.
Each stage addresses a distinct failure mode: without Stage~1, task RL lacks a whole-body scaffold and struggles to survive; without Stage~2, the policy lacks a reliable free-kick interaction primitive and converges to a poor local optimum which is only tracking the motions; without the Stage~3 planner, the kick timing fails to adapt; all leading to drops in performances.
Removing adaptive sampling or stablization phase severely disrupts the robustness and stability completing the shooting task, whose alive rate drops from 98.8\% to 32.8\% and 24.4\% accordingly, though while maintaining high shooting power.
Replacing our instant interaction reward with HDMI-style sparser ones drops contact accuracy performance, with free-kick success from 28.8\% to 2.5\%.

\section{Real-World Experiments}
\label{sec:realworld}

\subsection{Hardware Setup and Perception}
\begin{figure}[t]
  \centering
  \vspace{-20mm}  \includegraphics[width=\linewidth]{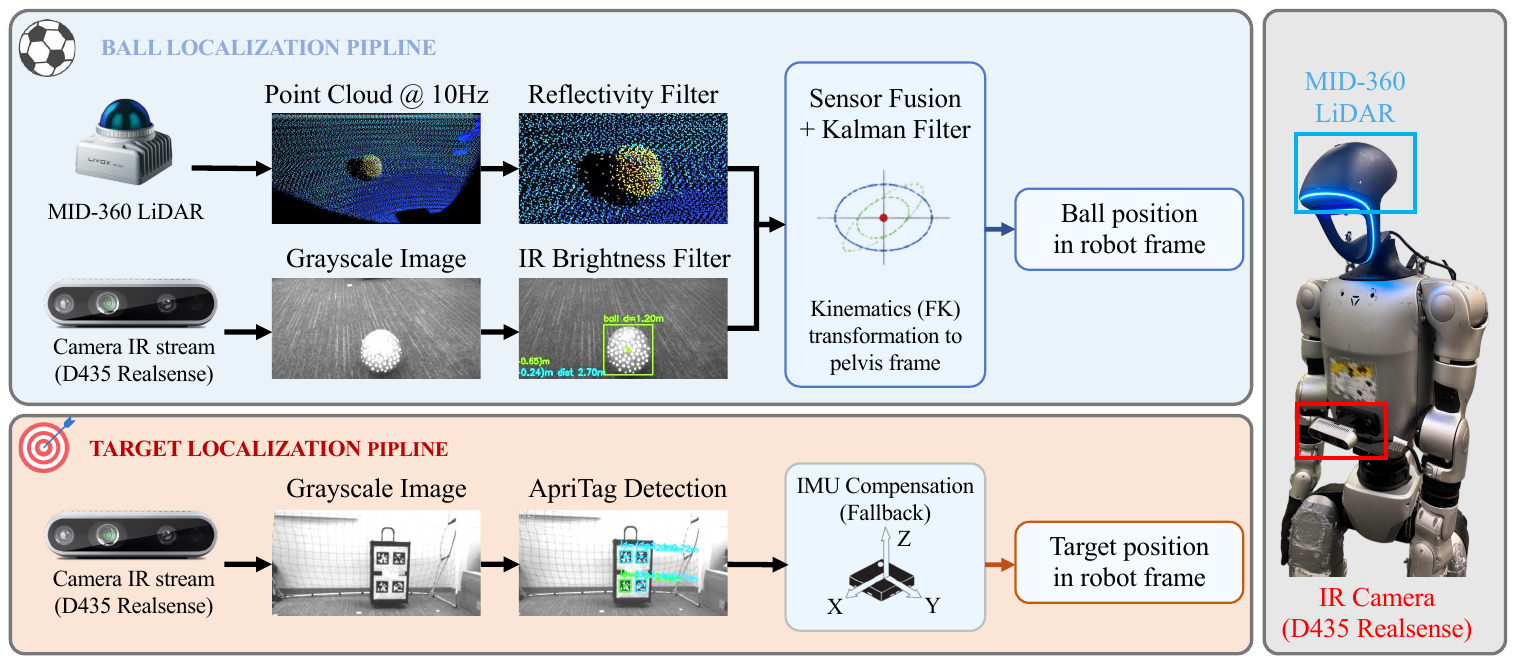}
  \vspace{-30mm}
  \caption{\textbf{Real-world perception stack.}
    A head-mounted MID-360 LiDAR tracks the retro-reflective ball at close range, while a chest-mounted IR camera extends ball detection beyond the LiDAR near-field.  Both estimates are fused by a constant-velocity Kalman filter and transformed into the pelvis frame for policy input.  The target board is localized by AprilTags from the same chest camera.}
  \label{fig:perception}
  \vspace{-4mm}
\end{figure}

Hardware experiments use a Unitree G1~\citep{unitreeg1} (29 DOF, 35\,kg) and a size-5 soccer ball on indoor floors and outdoor football fields.
The ONNX policy runs onboard at 50\,Hz over DDS, with no motion-capture or offboard state estimation.
A head-mounted Livox MID-360 and chest-mounted RealSense D435 localize the ball and AprilTag target (\cref{fig:perception}), so the trials exercise the full perception--control loop rather than a privileged-state replay. 
The same deployed controller is used for indoor calibration, point-target free-kicks, human-passed moving balls, and outdoor football-field demonstrations.

\begin{figure}[t]
  \centering
  \includegraphics[width=\linewidth]{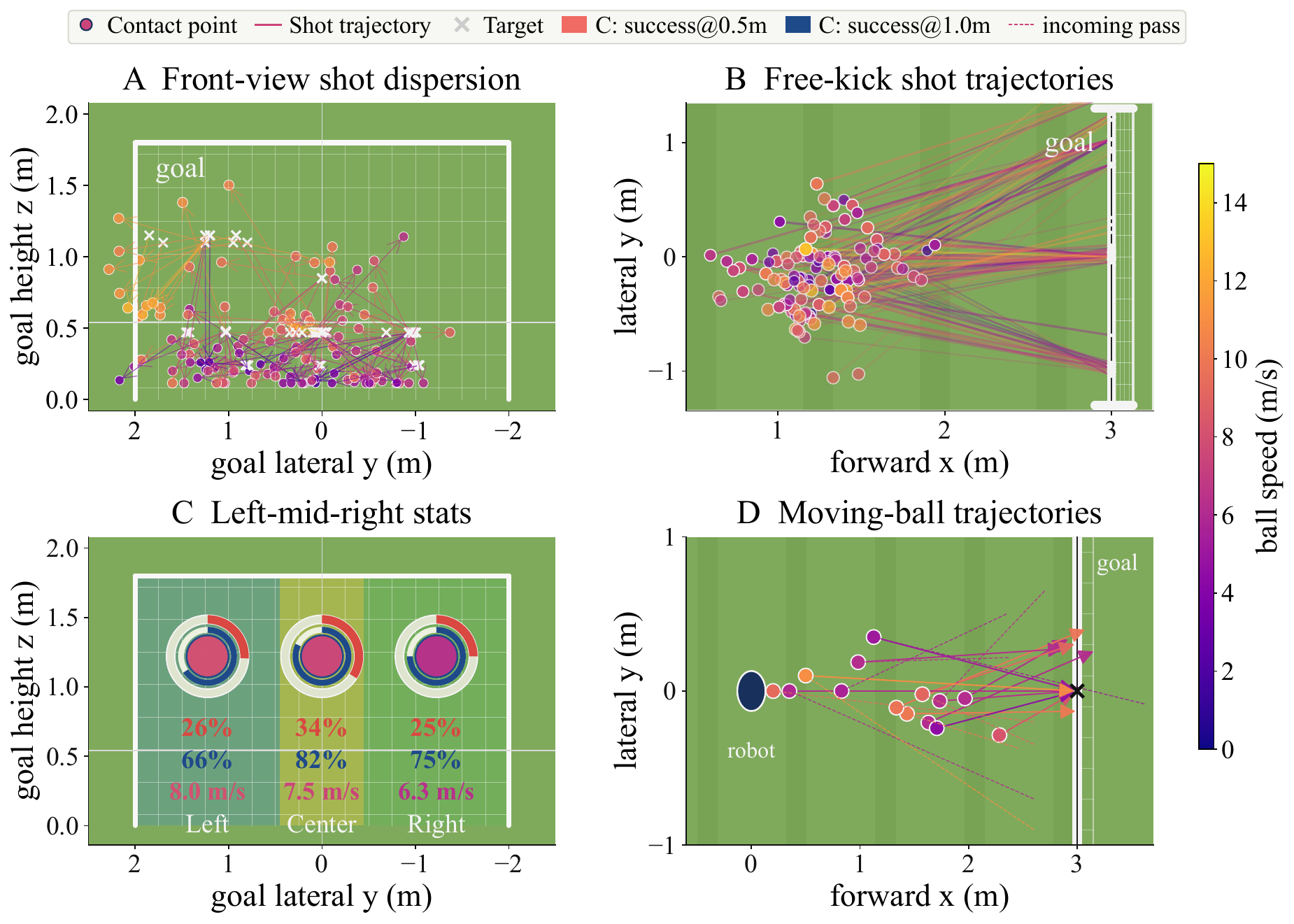}
  \caption{\textbf{Real-robot shooting overview.}
    \emph{A:} front-goal view shot dispersion; \emph{B:} free-kick shot trajectories; \emph{C:} per-region accuracy success rates and speed; \emph{D:} moving-ball shot trajectories.}
  \label{fig:real-static-kick}
\end{figure}

\subsection{Experimental Results}

We report two trial counts in \cref{tab:hardware}: $n_\text{total}$ counts all recorded attempts, while $n_\text{valid}$ counts attempts that successfully launch the ball and therefore have a measurable trajectory. Accuracy metrics, including the 0.5\,m and 1.0\,m success rates, shot error, and ball speed, are computed over $n_\text{valid}$ launches. Contact\% and Alive\% are reported over $n_\text{total}$ attempts to separately characterize end-to-end reliability.

\method{} transfers to hardware with reliable free-kick performance across target locations. Out of 136 recorded free-kick attempts, the robot launches 124 valid shots, corresponding to 91.2\% contact, and remains alive in all trials. Conditioned on valid launches, 31.5\% and 80.6\% of shots land within 0.5\,m and 1.0\,m of the target, respectively, with a mean radial error of $0.73$\,m and a mean ball speed of $7.42$\,m/s. The left, center, and right target breakdown shows that the policy has a natural right-footed bias: shots towards left-side has larger speed; center targets are the most accurate, while off-center targets remain close in 1.0\,m success, despite the increased task difficulty.

Moving-ball trials expose a different bottleneck. Human passes introduce timing variation and off-reference approach states, reducing contact reliability, but the kick remains accurate once a clean launch is produced. Among 27 moving-ball attempts, the robot launches 20 valid shots; conditioned on these launches, 30.0\% and 70.0\% land within 0.5\,m and 1.0\,m of the target, with $0.86$\,m mean shot error at $7.10$\,m/s. We observe in real-world experiment that the main degradation in the moving-ball setting comes from the sim-to-real gap of kick-timing trigger and imperfect incoming passes from human.

Compared to professional soccer players, \method{} reaches 13.10\,m/s peak ball speed, 59--71\% of reported professional open-play shot velocities~\citep{delacruz2026shotvelocity}, indicating high-impulse contact rather than conservative low-velocity strikes.

\begin{figure}[!t]
  \centering
  \newcommand{\panellab}[3]{\node[anchor=north west,font=\bfseries\footnotesize,text=white,
    fill=black,fill opacity=0.55,text opacity=1,inner sep=1.2pt] at (#1,#2) {#3};}
  \begin{tikzpicture}
    \node[anchor=south west,inner sep=0] (imgtop) {\includegraphics[width=0.9\linewidth]{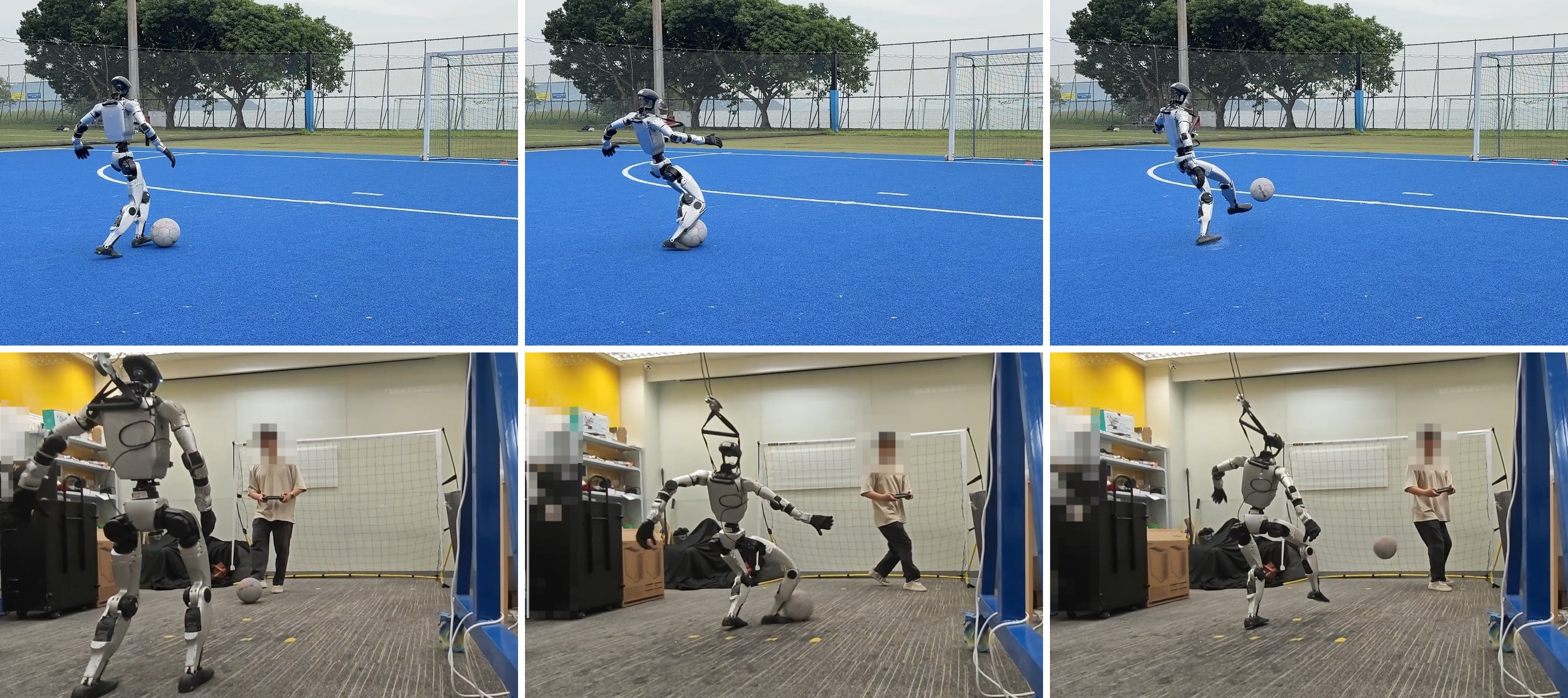}};
    \begin{scope}[x={(imgtop.south east)},y={(imgtop.north west)}]
      \panellab{0.004}{0.99}{(a)}
      \panellab{0.004}{0.49}{(b)}
    \end{scope}
  \end{tikzpicture}\\[-0.25em]
  \begin{tikzpicture}
    \node[anchor=south west,inner sep=0] (imgbot) {\includegraphics[width=0.9\linewidth]{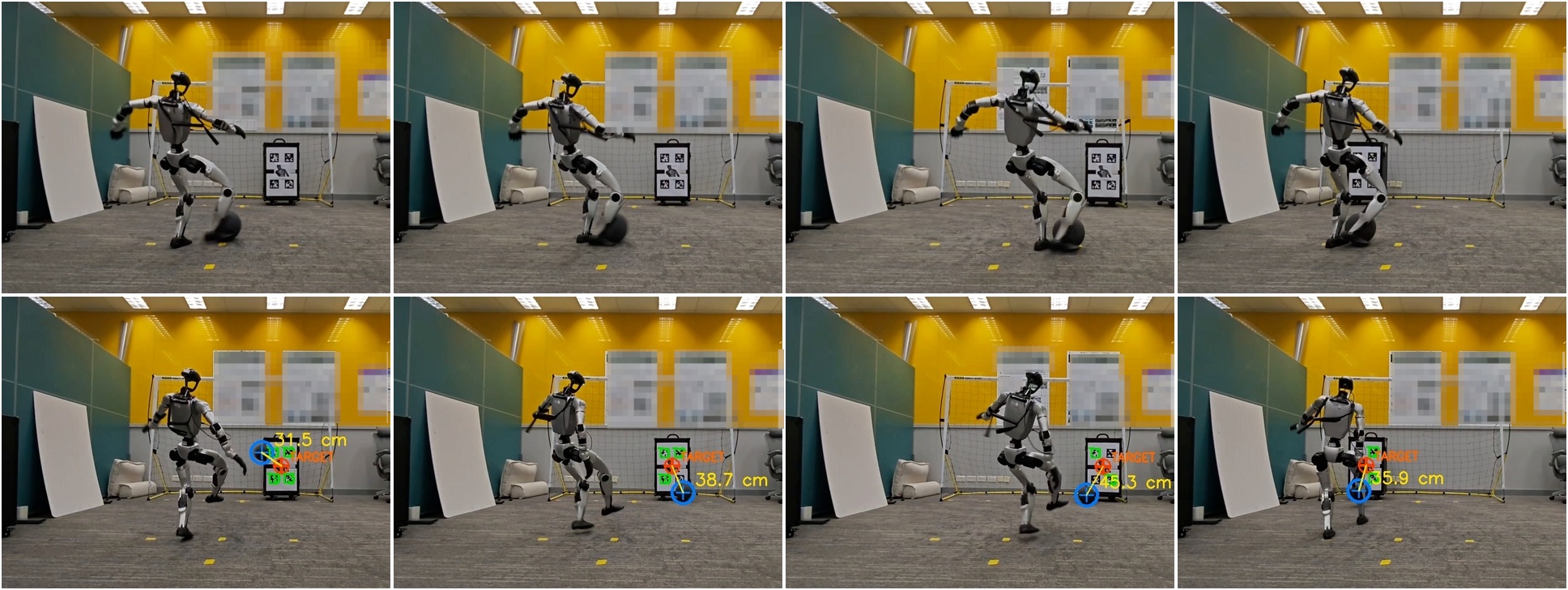}};
    \begin{scope}[x={(imgbot.south east)},y={(imgbot.north west)}]
      \panellab{0.004}{0.99}{(c)}
      \panellab{0.254}{0.99}{(d)}
      \panellab{0.504}{0.99}{(e)}
      \panellab{0.754}{0.99}{(f)}
    \end{scope}
  \end{tikzpicture}
  \caption{\textbf{Real-robot shooting showcase.}
    \textbf{(a)}~outdoor free-kick and \textbf{(b)}~moving-ball one-touch shooting are time-lapses read left-to-right.
    \textbf{(c)--(f)}~are four free-kick trials read top-to-bottom (contact frame above, landing outcome below), with landing errors of $31.5$, $38.7$, $45.3$, and $35.9$\,cm.}
  \label{fig:hardware}
  \vspace{-5mm}
\end{figure}

\begin{table}[t]
  \centering
  \caption{\textbf{Real-world shooting results.} Metrics follow \cref{tab:main}; moving-ball targets are random.}
  \label{tab:hardware}
  \footnotesize
  \resizebox{\linewidth}{!}{%
\begin{tabular}{@{}lcccccccc@{}}
  \toprule
  & $n_\text{total}$ & $n_\text{valid}$ & 0.5m suc. (\%) $\uparrow$ & 1.0m suc. (\%) $\uparrow$ & Shot err. (m) $\downarrow$ & $v_\text{ball}$ (m/s) $\uparrow$ & Contact (\%) $\uparrow$ & Alive (\%) $\uparrow$ \\
  \midrule
  \textit{Free-kick} \\
    \quad Overall & 136 & 124 & 31.5 & 80.6 & $0.73\pm0.39$ & $7.42\pm2.62$ & 91.2 & 100.0 \\
  \addlinespace[2pt]
    \quad\quad Left & 58 & 49 & 30.6 & 77.6 & $0.74\pm0.33$ & $7.97\pm2.98$ & 84.5 & 100.0 \\
    \quad\quad Center & 50 & 48 & 35.4 & 85.4 & $0.65\pm0.34$ & $7.47\pm2.30$ & 96.0 & 100.0 \\
    \quad\quad Right & 28 & 27 & 25.9 & 77.8 & $0.85\pm0.52$ & $6.35\pm2.06$ & 96.4 & 100.0 \\
  \addlinespace[4pt]
  \textit{Moving-ball} & 27 & 20 & 30.0 & 70.0 & $0.86\pm0.47$ & $7.10\pm2.40$ & 74.1 & 88.9 \\
  \bottomrule
\end{tabular}
  }
\end{table}


\section{Discussion and Conclusion}
\label{sec:conclusion}

We presented \method, a motion-guided curriculum RL framework for stable, accurate, and high-impulse humanoid interactions.
\method{} learns and demonstrates perceptive, powerful and accurate free-kicks and interactive moving-ball shots on a physical Unitree G1 with generalizability and stability. In real-world experiments, \method{} achieves 0.73\,m average target error from 3\,m, ball speeds up to 13.10\,m/s, and 74\% contact in human-passed moving-ball trials. These results suggest that motion-guided curriculum learning is effective for high-impulse humanoid interaction tasks requiring balance, aim, power, and contact timing.

\paragraph{Limitations and future work.}
\method{} currently relies on a single reference kick motion and a handcrafted high-level trigger policy for stage transitions and kick timing. This limits its ability to handle richer soccer scenarios that require multiple skills and autonomous decision making. Future work will investigate multi-skill motion priors and train high-level policies for skill selection, transition and timing in more complex cases. The perception module is also tailored to a retro-reflective ball; extending it to active egocentric vision to perceive natural ball is an important next step.


\acknowledgments{This work is in part supported by the JC STEM Lab of Autonomous Intelligent Systems funded by The Hong Kong Jockey Club Charities Trust; as well as the seed fund from Institutes of Data Science, The University of Hong Kong. We
gratefully acknowledge our robot carriers who help deliver the robots from lab to the outdoor playground: Ping Deng, Jing Wang and Lijia Yao. We also extend our thanks to Yuman Gao, Zike Yan, Ziye Wang, Jialong Zeng and Yucheng Huang.
}

\FloatBarrier

\bibliography{example}

@article{schulman2017proximal,
  title   = {Proximal Policy Optimization Algorithms},
  author  = {Schulman, John and Wolski, Filip and Dhariwal, Prafulla and Radford, Alec and Klimov, Oleg},
  journal = {arXiv preprint arXiv:1707.06347},
  year    = {2017}
}

@article{peng2018deepmimic,
  title     = {DeepMimic: Example-Guided Deep Reinforcement Learning of Physics-Based Character Skills},
  author    = {Peng, Xue Bin and Abbeel, Pieter and Levine, Sergey and van de Panne, Michiel},
  journal   = {ACM Transactions on Graphics (TOG)},
  volume    = {37},
  number    = {4},
  pages     = {1--14},
  year      = {2018}
}

@article{peng2021amp,
  title     = {AMP: Adversarial Motion Priors for Stylized Physics-Based Character Control},
  author    = {Peng, Xue Bin and Ma, Ze and Abbeel, Pieter and Levine, Sergey and Kanazawa, Angjoo},
  journal   = {ACM Transactions on Graphics (TOG)},
  volume    = {40},
  number    = {4},
  pages     = {1--20},
  year      = {2021}
}

@inproceedings{rudin2022learning,
  title     = {Learning to Walk in Minutes Using Massively Parallel Deep Reinforcement Learning},
  author    = {Rudin, Nikita and Hoeller, David and Reist, Philipp and Hutter, Marco},
  booktitle = {Conference on Robot Learning},
  pages     = {91--100},
  year      = {2022}
}

@article{mittal2023orbit,
  title   = {Orbit: A Unified Simulation Framework for Interactive Robot Learning Environments},
  author  = {Mittal, Mayank and Yu, Calvin and Yu, Qinxi and Liu, Jingzhou and Rudin, Nikita and Hoeller, David and Yuan, Jia Lin and Singh, Ritvik and Guo, Yunrong and Mazhar, Hammad and others},
  journal = {IEEE Robotics and Automation Letters},
  volume  = {8},
  pages   = {3740--3747},
  year    = {2023}
}

@inproceedings{tobin2017domain,
  title     = {Domain Randomization for Transferring Deep Neural Networks from Simulation to the Real World},
  author    = {Tobin, Josh and Fong, Rachel and Ray, Alex and Schneider, Jonas and Zaremba, Wojciech and Abbeel, Pieter},
  booktitle = {IEEE/RSJ International Conference on Intelligent Robots and Systems (IROS)},
  pages     = {23--30},
  year      = {2017}
}

@inproceedings{peng2018sim,
  title     = {Sim-to-Real Transfer of Robotic Control with Dynamics Randomization},
  author    = {Peng, Xue Bin and Andrychowicz, Marcin and Zaremba, Wojciech and Abbeel, Pieter},
  booktitle = {IEEE International Conference on Robotics and Automation (ICRA)},
  pages     = {3803--3810},
  year      = {2018}
}

@article{narvekar2020curriculum,
  title     = {Curriculum Learning for Reinforcement Learning Domains: A Framework and Survey},
  author    = {Narvekar, Sanmit and Peng, Bei and Leonetti, Matteo and Sinapov, Jivko and Taylor, Matthew E and Stone, Peter},
  journal   = {Journal of Machine Learning Research},
  volume    = {21},
  number    = {181},
  pages     = {1--50},
  year      = {2020}
}

@inproceedings{bengio2009curriculum,
  title     = {Curriculum Learning},
  author    = {Bengio, Yoshua and Louradour, J{\'e}r{\^o}me and Collobert, Ronan and Weston, Jason},
  booktitle = {Proceedings of the 26th Annual International Conference on Machine Learning},
  pages     = {41--48},
  year      = {2009}
}

@article{haarnoja2024learning,
  title     = {Learning Agile Soccer Skills for a Bipedal Robot with Deep Reinforcement Learning},
  author    = {Haarnoja, Tuomas and Moran, Ben and Lever, Guy and Huang, Sandy H and Tirumala, Dhayalan and Humplik, Jan and Wulfmeier, Markus and Tunyasuvunakool, Saran and Siegel, Noah Y and Hafner, Roland and others},
  journal   = {Science Robotics},
  volume    = {9},
  number    = {89},
  pages     = {eadi8022},
  year      = {2024}
}

@misc{jocher2024yolo11,
  title        = {Ultralytics {YOLO11}},
  author       = {Jocher, Glenn and Qiu, Jing},
  howpublished = {\url{https://docs.ultralytics.com/models/yolo11/}},
  year         = {2024}
}

@inproceedings{todorov2012mujoco,
  title     = {MuJoCo: A Physics Engine for Model-Based Control},
  author    = {Todorov, Emanuel and Erez, Tom and Tassa, Yuval},
  booktitle = {IEEE/RSJ International Conference on Intelligent Robots and Systems (IROS)},
  pages     = {5026--5033},
  year      = {2012}
}

@misc{unitreeg1,
  title        = {Unitree G1 Humanoid Robot},
  author       = {{Unitree Robotics}},
  howpublished = {\url{https://www.unitree.com/g1}},
  year         = {2024}
}

@inproceedings{makoviychuk2021isaac,
  title     = {Isaac Gym: High Performance GPU-Based Physics Simulation For Robot Learning},
  author    = {Makoviychuk, Viktor and Wawrzyniak, Lukasz and Guo, Yunrong and Lu, Michelle and Storey, Kier and Macklin, Miles and Hoeller, David and Rudin, Nikita and Allshire, Arthur and Handa, Ankur and others},
  booktitle = {Advances in Neural Information Processing Systems},
  year      = {2021}
}

@article{wang2026humanx,
  title   = {{HumanX}: Toward Agile and Generalizable Humanoid Interaction Skills from Human Videos},
  author  = {Wang, Yinhuai and Zhao, Qihan and Lau, Yuen Fui and Yu, Runyi and Tsui, Hok Wai and Chen, Qifeng and Wang, Jingbo and Pang, Jiangmiao and Tan, Ping},
  journal = {arXiv preprint arXiv:2602.02473},
  year    = {2026}
}

@article{kong2026paid,
  title   = {Learning Soccer Skills for Humanoid Robots: A Progressive Perception-Action Framework},
  author  = {Kong, Jipeng and Liu, Xinzhe and Lin, Yuhang and Han, Jinrui and Schwertfeger, S{\"o}ren and Bai, Chenjia and Li, Xuelong},
  journal = {arXiv preprint arXiv:2602.05310},
  year    = {2026}
}

@article{liao2025beyondmimic,
  title   = {BeyondMimic: From Motion Tracking to Versatile Humanoid Control via Guided Diffusion},
  author  = {Liao, Qiayuan and Truong, Takara E. and Huang, Xiaoyu and Gao, Yuman and Tevet, Guy and Sreenath, Koushil and Liu, C. Karen},
  journal = {arXiv preprint arXiv:2508.08241},
  year    = {2025}
}

@article{zhao2025resmimic,
  title   = {ResMimic: From General Motion Tracking to Humanoid Whole-body Loco-Manipulation via Residual Learning},
  author  = {Zhao, Siheng and Ze, Yanjie and Wang, Yue and Liu, C. Karen and Abbeel, Pieter and Shi, Guanya and Duan, Rocky},
  journal = {arXiv preprint arXiv:2510.05070},
  year    = {2025}
}

@article{araujo2025gmr,
  title   = {Retargeting Matters: General Motion Retargeting for Humanoid Motion Tracking},
  author  = {Ara{\'u}jo, Jo{\~a}o Pedro and Ze, Yanjie and Xu, Pei and Wu, Jiajun and Liu, C. Karen},
  journal = {arXiv preprint arXiv:2510.02252},
  year    = {2025}
}

@article{wang2025visiondriven,
  title   = {Learning Vision-Driven Reactive Soccer Skills for Humanoid Robots},
  author  = {Wang, Yushi and Luo, Changsheng and Chen, Penghui and Liu, Jianran and Sun, Weijian and Guo, Tong and Yang, Kechang and Hu, Biao and Zhang, Yangang and Zhao, Mingguo},
  journal = {arXiv preprint arXiv:2511.03996},
  year    = {2025}
}

@article{delacruz2026shotvelocity,
  title   = {The Role of Shot Velocity in Advanced Post-Shot Metrics: Evidence from the {UEFA} European Football Championships},
  author  = {De-la-Cruz-Torres, Blanca and Ruiz-de-Alarc{\'o}n-Quintero, Anselmo and Navarro-Castro, Miguel},
  journal = {Data},
  volume  = {11},
  number  = {2},
  pages   = {39},
  year    = {2026},
  doi     = {10.3390/data11020039}
}

@inproceedings{li2025kicktraj,
  title     = {Like Playing a Video Game: Spatial-Temporal Optimization of Foot Trajectories for Controlled Football Kicking in Bipedal Robots},
  author    = {Li, Wanyue and Ma, Ji and Lu, Minghao and Lu, Peng},
  booktitle = {IEEE/RSJ International Conference on Intelligent Robots and Systems (IROS)},
  pages     = {3565--3572},
  year      = {2025},
  doi       = {10.1109/IROS60139.2025.11246655},
  note      = {arXiv:2510.01843}
}

@inproceedings{xu2026striker,
  title     = {Learning Agile Striker Skills for Humanoid Soccer Robots from Noisy Sensory Input},
  author    = {Xu, Zifan and Seo, Myoungkyu and Lee, Dongmyeong and Fu, Hao and Hu, Jiaheng and Cui, Jiaxun and Jiang, Yuqian and Wang, Zhihan and Brund, Anastasiia and Biswas, Joydeep and Stone, Peter},
  booktitle = {IEEE International Conference on Robotics and Automation (ICRA)},
  year      = {2026},
  note      = {arXiv:2512.06571}
}

@article{ren2025goalkeeper,
  title   = {Humanoid Goalkeeper: Learning from Position Conditioned Task-Motion Constraints},
  author  = {Ren, Junli and Long, Junfeng and Huang, Tao and Wang, Huayi and Wang, Zirui and Jia, Feiyu and Zhang, Wentao and Wang, Jingbo and Luo, Ping and Pang, Jiangmiao},
  journal = {arXiv preprint arXiv:2510.18002},
  year    = {2025}
}

@article{su2025hitter,
  title   = {{HITTER}: A {HumanoId} Table {TEnnis} Robot via Hierarchical Planning and Learning},
  author  = {Su, Zhi and Zhang, Bike and Rahmanian, Nima and Gao, Yuman and Liao, Qiayuan and Regan, Caitlin and Sreenath, Koushil and Sastry, S. Shankar},
  journal = {arXiv preprint arXiv:2508.21043},
  year    = {2025}
}

@article{ma2025badminton,
  title   = {Learning coordinated badminton skills for legged manipulators},
  author  = {Ma, Yuntao and Cramariuc, Andrei and Farshidian, Farbod and Hutter, Marco},
  journal = {Science Robotics},
  volume  = {10},
  number  = {102},
  pages   = {eadu3922},
  year    = {2025},
  doi     = {10.1126/scirobotics.adu3922}
}

@article{weng2025hdmi,
  title   = {{HDMI}: Learning Interactive Humanoid Whole-Body Control from Human Videos},
  author  = {Weng, Haoyang and Li, Yitang and Sobanbabu, Nikhil and Wang, Zihan and Luo, Zhengyi and He, Tairan and Ramanan, Deva and Shi, Guanya},
  journal = {arXiv preprint arXiv:2509.16757},
  year    = {2025}
}

@article{kalaria2025dreamcontrol,
  title   = {{DreamControl}: Human-Inspired Whole-Body Humanoid Control for Scene Interaction via Guided Diffusion},
  author  = {Kalaria, Dvij and Harithas, Sudarshan S and Katara, Pushkal and Kwak, Sangkyung and Bhagat, Sarthak and Sastry, Shankar and Sridhar, Srinath and Vemprala, Sai and Kapoor, Ashish and Huang, Jonathan Chung-Kuan},
  journal = {arXiv preprint arXiv:2509.14353},
  year    = {2025}
}

@inproceedings{zhang2026falcon,
  title     = {{FALCON}: Learning Force-Adaptive Humanoid Loco-Manipulation},
  author    = {Zhang, Yuanhang and Yuan, Yifu and Gurunath, Prajwal and Gupta, Ishita and Omidshafiei, Shayegan and Agha-mohammadi, Ali-akbar and Vazquez-Chanlatte, Marcell and Pedersen, Liam and He, Tairan and Shi, Guanya},
  booktitle = {Learning for Dynamics and Control Conference (L4DC)},
  year      = {2026},
  note      = {arXiv:2505.06776}
}

@inproceedings{su2025quadruped,
  title     = {Toward Real-World Cooperative and Competitive Soccer with Quadrupedal Robot Teams},
  author    = {Su, Zhi and Gao, Yuman and Lukas, Emily and Li, Yunfei and Cai, Jiaze and Tulbah, Faris and Gao, Fei and Yu, Chao and Li, Zhongyu and Wu, Yi and Sreenath, Koushil},
  booktitle = {Conference on Robot Learning (CoRL)},
  year      = {2025},
  note      = {arXiv:2505.13834}
}

@inproceedings{xu2025intermimic,
  title     = {{InterMimic}: Towards Universal Whole-Body Control for Physics-Based Human-Object Interactions},
  author    = {Xu, Sirui and Ling, Hung Yu and Wang, Yu-Xiong and Gui, Liang-Yan},
  booktitle = {Proceedings of the IEEE/CVF Conference on Computer Vision and Pattern Recognition (CVPR)},
  month     = {June},
  year      = {2025},
  pages     = {12266--12277},
  doi       = {10.1109/CVPR52734.2025.01145},
  note      = {Highlight Paper}
}

@inproceedings{shen2024gvhmr,
  title     = {World-Grounded Human Motion Recovery via Gravity-View Coordinates},
  author    = {Shen, Zehong and Pi, Huaijin and Xia, Yan and Cen, Zhi and Peng, Sida and Hu, Zechen and Bao, Hujun and Hu, Ruizhen and Zhou, Xiaowei},
  booktitle = {SIGGRAPH Asia 2024 Conference Papers},
  year      = {2024},
  doi       = {10.1145/3680528.3687565}
}

\clearpage
\appendix
\counterwithin{figure}{section}
\counterwithin{table}{section}
\crefname{section}{Appendix}{Appendices}
\Crefname{section}{Appendix}{Appendices}

\section{Observation Space: Per-Dimension Breakdown}
\label{app:obs}

\begin{table}[h]
  \centering
  \caption{Policy observation vector (547 dims). $^\dagger$In Stage~3 the anchor reference is replaced by a 9-dim heuristic locomotion command (target body velocity/heading), keeping the total at 547. The critic additionally receives privileged, noise-free ground-truth body positions/orientations/velocities.}
  \label{tab:obs}
  \begin{tabular}{llc}
    \toprule
    Group & Component & Dim. \\
    \midrule
    \multirow{2}{*}{Motion reference}
      & reference joint positions, $\mathbf{q}_\text{ref}$ (29 DOF)    & 29 \\
      & reference joint velocities, $\dot{\mathbf{q}}_\text{ref}$ (29 DOF) & 29 \\
    \midrule
    \multirow{2}{*}{Anchor reference$^\dagger$}
      & anchor (torso) position in body frame, $\mathbf{p}_\text{anc}$ & 3 \\
      & anchor orientation (6D rotation), $\mathbf{R}_\text{anc}$       & 6 \\
    \midrule
    \multirow{4}{*}{Proprioception ($\times$5 history)}
      & base angular velocity                                          & 15 \\
      & joint positions (29 DOF)                                       & 145 \\
      & joint velocities (29 DOF)                                      & 145 \\
      & previous actions (29 DOF)                                      & 145 \\
    \midrule
    \multirow{3}{*}{Exteroception ($\times$5 history)}
      & ball position in robot frame                                   & 15 \\
      & target position in robot frame                                 & 15 \\
    \midrule
    & \textbf{Total}                                                   & \textbf{547} \\
    \bottomrule
  \end{tabular}
\end{table}

\section{Reward Terms and Weights}
\label{app:reward}

\cref{tab:rewards} lists every reward term.
Motion-tracking terms use an exponential kernel $\exp(-\|\mathbf{e}\|^2/\sigma^2)$ on the indicated error $\mathbf{e}$ with the listed base standard deviation $\sigma_0$; in Stages~2--3 the effective standard deviation is $\sigma_0/\alpha$ with $\alpha{=}2$, and in Stage~3 each motion-tracking term is additionally multiplied by the proximity-based relaxation factor $g_i(d_\text{ball})$ (\cref{tab:prox_relax}).
Task terms are scaled by the goal weight $w_g$ ($0$ in Stage~1, $1$ thereafter); regularization terms are scaled by $w_r$ ($0.2$ in Stages~1--2, $0.25$ in Stage~3).
The 14 tracked bodies are the pelvis, the bilateral hip-roll, knee, and ankle-roll links, the torso, and the bilateral shoulder-roll, elbow, and wrist-yaw links.

\begin{table}[h]
  \centering
  \caption{Reward terms. Motion-tracking terms use an exp.\ kernel $\exp(-\|\mathbf e\|^2/\sigma^2)$; $\sigma_0$ is the base std, with $\sigma{=}\sigma_0/\alpha$ ($\alpha{=}2$) in Stages~2--3 and a proximity-based relaxation factor $g_i(d_\text{ball})$ in Stage~3 (\cref{tab:prox_relax}). $w_g$: goal weight ($0$/$1$); $w_r$: regularization weight ($0.2$/$0.25$); per-stage scale shown as St.\,1--2\,/\,St.\,3.}
  \label{tab:rewards}
  \footnotesize
  \resizebox{\linewidth}{!}{%
  \begin{tabular}{p{5.0cm} l p{2.6cm}}
    \toprule
    Term & Quantity / form & Weight \\
    \midrule
    \multicolumn{3}{l}{\emph{Motion tracking}}\\
    anchor position ($\sigma_0{=}0.3$\,m)        & torso position vs.\ ref.            & $0.1$ \\
    anchor orientation ($\sigma_0{=}0.4$\,rad)   & torso orientation vs.\ ref.         & $0.5$ \\
    body positions ($\sigma_0{=}0.3$\,m)         & 14 relative body positions          & $1.0$ \\
    body orientations ($\sigma_0{=}0.4$\,rad)    & 14 body orientations                & $1.0$ \\
    body linear vel.\ ($\sigma_0{=}1.0$\,m/s)    & 14 body linear velocities           & $1.0$ \\
    body angular vel.\ ($\sigma_0{=}\pi$\,rad/s) & 14 body angular velocities          & $1.0$ \\
    feet linear vel.\ ($\sigma_0{=}1.0$\,m/s)    & 2 feet linear velocities            & $1.0$ \\
    \midrule
    \multicolumn{3}{l}{\emph{Task} ($\times w_g$)}\\
    \texttt{error\_ball\_to\_target}             & exp; $\min_t\|\mathbf p_b{-}\mathbf p_t\|$ ($\sigma{=}0.5/\alpha$); $=1$ within thr. & $5 w_g$ \\
    \texttt{predicted\_error\_ball\_to\_target}  & exp; $\|\hat{\mathbf p}{-}\mathbf p_t\|$ (Eq.\,3, $\sigma{=}1/\alpha$) & $10 w_g$ \\
    \texttt{goal\_reward\_burst}                 & constant for 10 steps on success (delay 100) & $300 w_g$ \\
    \texttt{robot\_ball\_contact}                & ankle--$\mathbf p_\text{exec}$ proximity $+$ contact force, buffered & $2 w_g$ \\
    contact orientation                          & ankle velocity vs.\ ball$\to$target & $1 w_g$ \\
    ankle--ball / torso--ball distance           & $d_\text{ball}$ / torso--ball dist. & $1 w_g$ / $1.5 w_g$ \\
    \texttt{ball\_velocity}                      & sat.\ Lorentzian on $\|\mathbf v_b\|$ post-contact & $0.5 w_g$ \\
    alive                                        & constant while upright              & $0.5 w_g$ \\
    \midrule
    \multicolumn{3}{l}{\emph{Regularization} ($\times w_r$)}\\
    action rate                                  & $-\|\Delta\mathbf a\|^2$            & $-0.1$ (1.0/1.5) \\
    action smoothness                            & $-\|\Delta^2\mathbf a\|^2$          & $-10^{-5}$ (1.0/2.0) \\
    feet slip                                    & $-\|\mathbf v_\text{foot}\|^2$ in stance & $-10^{-2}$ \\
    no-fly                                       & $-\mathbb 1[\text{both feet airborne}]$ & $-0.2$ (1.0/1.25) \\
    joint velocity                               & $-\|\dot{\mathbf q}\|^2$            & $-10^{-5}$ \\
    joint torque                                 & $-\|\boldsymbol\tau\|^2$            & $-2{\times}10^{-7}$ \\
    joint limit (always active)                  & $-\mathbb 1[\text{violation}]$      & $-10$ (1.0/1.5) \\
    undesired contacts                           & contact off ankles/wrists           & $-0.1$ \\
    feet contact time                            & smooth foot-contact duration        & $-0.05$ (1.0/1.2) \\
    \bottomrule
  \end{tabular}}
\end{table}


\begin{table}[h]
  \centering
  \caption{\textbf{Three-stage curriculum.} Ball spawns relative to
  $(1.0, -0.2, 0.115)$\,m; $^\dagger$kick/loco mode. Full observation vector in \cref{tab:obs}.}
  \label{tab:curriculum}
  \begin{tabular}{lccc}
    \toprule
    & \textbf{S1} & \textbf{S2} & \textbf{S3} \\
    & Tracking & Adapt & Generalize \\
    \midrule
    $w_{\text{task}}$   & 0.0 & 0.8 & 1.0 \\
    $w_{\text{reg}}$    & 0.0 & 0.2 & 0.25 \\
    $w_{\text{motion}}$ & 1.0 & 1.0 & 1.0\,/\,0.1$^\dagger$ \\
    Ball spawn XY (m)   & --- & $\pm$0.75 & $\pm$1.0 \\
    Ball vel.\ XY (m/s) & 0 & 0.1\,m/s & 5\,m/s \\
    \midrule
    \shortstack[l]{EEF $\Delta p_z$\\term.\ (m)~\citep{liao2025beyondmimic}}
      & 0.25 & 0.5 & 0.5 \\
    \bottomrule
  \end{tabular}
\end{table}

\paragraph{Instant Interaction Reward factors.}
The compact reward in \cref{eq:interact} uses
\begin{align}
  R_\text{contact} &= \exp\!\Bigl(-\tfrac{[\|\mathbf{p}_\text{foot}{-}\mathbf{p}_\text{ball}\|-r_b]_+}{\sigma_f}\Bigr), &
  R_\text{goal} &= \exp\!\Bigl(-\tfrac{[\hat{d}-\frac{\sigma_g}{2}]_+^2}{(2\sigma_g)^2}\Bigr), \label{eq:factors}\\
  R_\text{vel} &= 1-\exp\!\Bigl(-\tfrac{[\|\mathbf{v}_\text{ball}\|-v_\text{min}]_+^2}{10}\Bigr), &
  R_\text{force} &= \min\!\Bigl(\tfrac{f}{f_\text{thresh}},1\Bigr), \notag
\end{align}
where $\hat{d}=\min_{t'\le t}d_\text{ball-target}(t')$ and $[\cdot]_+=\max(\cdot,0)$.

\paragraph{Densified Shooting Reward.}
Shot placement is fully determined at the moment of impact, yet the episode-minimum ball--target distance reward yields signal only when the ball eventually reaches the goal line.
To close this credit-assignment gap, at every post-contact step we ballistically extrapolate the ball's current velocity under gravity to its goal-line crossing:
\begin{align}
  \label{eq:densified}
  t^* &= \frac{p_{t,y}-p_{b,y}}{v_{b,y}+\epsilon}, \qquad
  \hat{\mathbf{p}} = \mathbf{p}_b + \mathbf{v}_b\,t^* - \tfrac{1}{2}g\,(t^*)^2\hat{\mathbf{z}},\\
  r_\text{densified} &= \begin{cases}
    \exp\!\bigl(-\|\hat{\mathbf{p}}-\mathbf{p}_t\|^2/\sigma^2\bigr) & t^*>0\\[2pt]
    0 & \text{else,}
  \end{cases} \nonumber
\end{align}
with $\hat{p}_z$ clamped to $\ge$ the ball radius.
The condition $t^*>0$ activates the reward only when the ball moves toward the goal; once the ball crosses the goal line the reward switches to the actual landing position.

\paragraph{Proximity-based tracking relaxation.}
\label{sec:prox_relax}
In Stage~3, each motion-tracking term is multiplied by a distance-conditioned relaxation factor:
\begin{equation}
  \label{eq:prox_relax}
  g_i(d) \;=\; \mu_i + (1-\mu_i)\cdot\mathrm{clamp}\!\left(\tfrac{d-d_\text{near}}{d_\text{far}-d_\text{near}},\,0,\,1\right),
  \qquad d_\text{near}=0.35\text{\,m},\;\; d_\text{far}=1.0\text{\,m},
\end{equation}
with per-term near-ball scales $\mu_i$ given in \cref{tab:prox_relax}.
Feet linear velocity is suppressed to $\mu{=}0.05$, giving the policy near-total freedom in ankle placement during contact; body orientation is only reduced to $\mu{=}0.6$, preserving balance during the lean.
Unlike globally zeroing a tracking weight, this design retains full reference pressure during approach and relaxes only the terms where the task objective genuinely conflicts with motion tracking.

\begin{table}[h]
  \centering
  \caption{\textbf{Proximity-based tracking relaxation scales} $\mu_i$
  (Stage~3, $d_\text{ball}\!\le\!d_\text{near}{=}0.35$\,m).}
  \label{tab:prox_relax}
  \small
  \resizebox{\linewidth}{!}{%
  \begin{tabular}{lccccccc}
    \toprule
    Term & anchor pos. & anchor ori. & body pos. & body ori. & body lin.\,vel. & body ang.\,vel. & feet lin.\,vel. \\
    \midrule
    $\mu_i$ & 0.10 & 0.60 & 0.20 & 0.60 & 0.20 & 0.30 & 0.05 \\
    \bottomrule
  \end{tabular}}
\end{table}

\section{Hyperparameters and Domain Randomization}
\label{app:hparams}

\paragraph{PPO (RSL-RL).}
We optimize the policy with PPO~\citep{schulman2017proximal} using the massively parallel RSL-RL implementation~\citep{rudin2022learning}.
Clip ratio $0.2$; entropy coefficient $0.001$; value-loss coefficient $1.0$ with clipped value loss; $5$ learning epochs and $4$ minibatches per update; $24$ environment steps per rollout; adaptive learning rate (initial $2\times10^{-5}$) with KL target $0.01$; discount $\gamma{=}0.99$; GAE $\lambda{=}0.95$; gradient-norm clip $1.0$; up to $10^5$ iterations per stage with checkpoints every $10^3$.
Actor and critic are $3$-layer MLPs ($512{\to}256{\to}128$, ELU) with empirical observation normalization; initial action noise std $1.0$.
The critic uses privileged, noise-free observations; the actor uses the noised observation of \cref{app:obs}.

\paragraph{Heuristic planner.}
Prediction horizon $T_h{=}0.4\,\mathrm{s}$; proximity radius $r_\mathrm{thr}{=}0.25\,\mathrm{m}$; maximum kickable height $h_\mathrm{max}{=}0.75\,\mathrm{m}$; locomotion–kick boundary frame $\tau_\mathrm{kick}{=}240$; adaptive sampling window $\pm10$ frames; foot-strike offset $\mathbf{p}_\mathrm{offset}{=}(0.15,\,-0.1)\,\mathrm{m}$.
The desired anchor command is $\mathbf{p}_\text{des}=\mathbf{p}_{\text{anc},\perp}+\tfrac{1}{2}\min(\|\tilde{\mathbf{p}}_b-\mathbf{p}_{\text{anc},\perp}\|,2)\hat{\mathbf{d}}_b$ and $\mathbf{q}_\text{des}=\mathrm{norm}([1+\mathbf{e}_x{\cdot}\hat{\mathbf{d}}_b,\mathbf{e}_x{\times}\hat{\mathbf{d}}_b])$, where $\tilde{\mathbf{p}}_b{=}\mathbf{p}_{b,\perp}{+}\mathbf{v}_{b,\perp}T_h$.
The predicted closest-approach distance is $d_\text{min}=\|\Delta\mathbf{p}+t^*\mathbf{v}_{b,\perp}^\text{body}\|$, with $\Delta\mathbf{p}=\mathbf{p}_{b,\perp}^\text{body}-\mathbf{p}_\text{offset}$ and $t^*=\mathrm{clamp}(-\Delta\mathbf{p}{\cdot}\mathbf{v}_{b,\perp}^\text{body}/\|\mathbf{v}_{b,\perp}^\text{body}\|^2,0,T_h)$.

\paragraph{Simulation.}
$4096$ parallel environments on a $10\times10$\,m grid in GPU-accelerated simulation~\citep{makoviychuk2021isaac}; PhysX GPU solver with $8$ position and $4$ velocity iterations; physics step $5$\,ms ($200$\,Hz) with $4\times$ decimation ($50$\,Hz control); episodes of $10$\,s ($500$ steps).
The ball is a rigid sphere (radius $0.115$\,m, mass $\approx0.41$\,kg, restitution $0.95$) with contact sensors; the active ankle-roll link has a contact sensor for kick detection.

\paragraph{Domain randomization.}
Following standard sim-to-real randomization practice~\citep{tobin2017domain,peng2018sim}, per episode: robot collision material: static friction $\mathcal U[0.3,1.6]$, dynamic friction $\mathcal U[0.3,1.2]$, restitution $\mathcal U[0,0.5]$; joint default-position offsets $\pm0.01$\,rad; torso center-of-mass offset $\pm0.025$\,m (fore--aft) and $\pm0.05$\,m (lateral/vertical); actuator command delay of a few control steps.
At $1$--$3$\,s intervals a random base push is applied: linear velocity $\pm0.5$\,m/s (horizontal), $\pm0.2$\,m/s (vertical); angular velocity $\pm0.52$\,rad/s (roll/pitch), $\pm0.78$\,rad/s (yaw).
Ball spawn position and (in Stage~3) initial velocity are randomized as in \cref{tab:curriculum}.

\section{Deployment Details}
\label{app:deploy}

\begin{table}[h]
  \centering\footnotesize
  \caption{\textbf{Perception modules.}
    \emph{Fast ball}: robust to 15\,m/s motion blur.
    \emph{Onboard}: no external infra.}
  \label{tab:perception}
  \begin{tabular}{lcccc}
    \toprule
    Module & Range & Lat. (ms) & Fast ball & Onboard \\
    \midrule
    MoCap (Vicon)           & room   & $<$5   & \cmark & \xmark \\
    RGB (D435 color)        & 6\,m   & 30--60 & \xmark & \cmark \\
    Depth (D435 depth)      & 3\,m   & 30--60 & {--}   & \cmark \\
    \textbf{LiDAR (MID-360)} & \textbf{40\,m} & \textbf{10--20} & \cmark & \cmark \\
    \bottomrule
  \end{tabular}
\end{table}

\paragraph{Perception.}
\emph{Ball (LiDAR, near range).} A Livox MID-360 is rigidly mounted on the head with a downward pitch so that its dense returns cover the floor from a few decimetres to about two metres in front of the robot.
Each $\sim$10\,Hz point cloud is filtered by (i)~reflectivity (the retro-reflective soccer ball returns are far brighter than surroundings), (ii)~a distance gate, and (iii)~a region of interest; a sphere of known radius is fit (least-squares with outlier rejection) to the surviving points, the center is smoothed by a constant-velocity Kalman filter, transformed to the pelvis frame, and published on \texttt{rt/ball\_state}.
\emph{Ball (IR$+$grayscale camera, far range).} Beyond the LiDAR coverage we localize the ball from the chest-mounted RealSense D435 \emph{infrared} (grayscale) stream rather than the RGB image: the retro-reflective ball is a bright blob in the IR image, segmented by intensity thresholding plus a known-radius circularity check, then back-projected and fused into the same Kalman filter as the LiDAR estimate.
Keying on retro-reflective brightness avoids the failure modes of the appearance-based detectors we also tried on the D435 RGB stream---a YOLO11m~\citep{jocher2024yolo11} TensorRT detector and an HSV$+$MOG2 background-subtraction detector---namely HSV color segmentation being easily disturbed by lighting and background, and the YOLO detector losing the ball under fast-motion blur.
Robust long-range tracking of a fast moving ball remains part of the planned extension to ball-stopping and interception.
\emph{Target (camera).} An AprilTag detector on the RealSense D435 localizes the goal board; alternatively a fixed coordinate is used.
Detections older than $300$\,ms are flagged invalid.

\paragraph{Joint-index remapping.}
Isaac Lab enumerates the 29 DOF in breadth-first order over the kinematic tree, whereas the Unitree SDK and the MuJoCo~\citep{todorov2012mujoco} model use depth-first order.
A fixed permutation (and its inverse) is applied when packing the observation and when mapping policy actions to motor commands.
Getting this permutation wrong silently produces a plausible-looking but non-functional policy, so we list it explicitly in the released code.

\paragraph{Initialization and control loop.}
On entry to the policy state, the five-step observation history buffers are zero-filled and a $10$-step linear warmup blends the robot's current joint configuration $\mathbf q_0$ into the first policy output $\mathbf a_1$: $\mathbf q_\text{target}^{(t)} = (1-t/10)\,\mathbf q_0 + (t/10)\,\mathbf a_1$ for $t=1,\dots,10$.
The control loop runs at $50$\,Hz: it reads \texttt{rt/lowstate} (joint state and IMU, published at $500$\,Hz by the firmware) and \texttt{rt/ball\_state}/\texttt{rt/target\_state}, runs ONNX inference, and publishes \texttt{rt/lowcmd} (per-joint position target with PD gains, CRC-checked).
The \method{} policy is one state of an on-robot finite-state machine; an immediate damping (passive) mode is bound to a dedicated button for safety.

\section{Additional Simulation Results}
\label{app:sim-extra}

\begin{figure}[h]
  \centering
  \includegraphics[width=\linewidth]{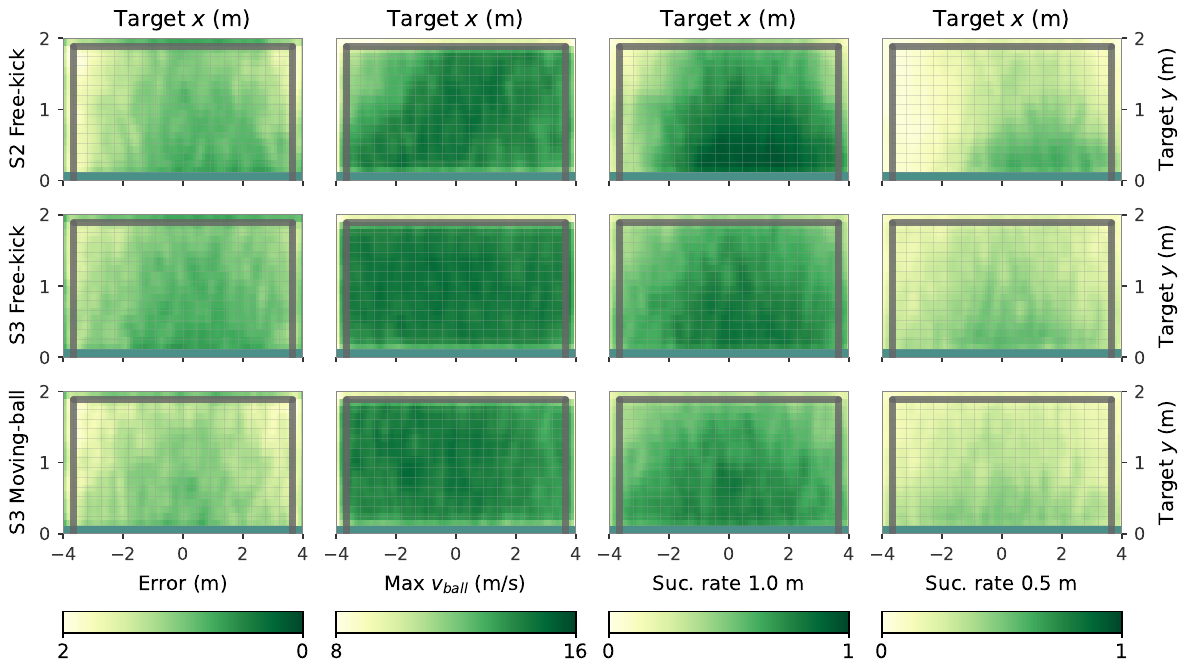}
  \caption{\textbf{Target-conditioned shooting heatmaps.}
    Rows correspond to Stage~2 free-kick, Stage~3 free-kick, and Stage~3 moving-ball shooting.
    Each heatmap visualizes one shooting metric over goal-plane target locations: shot error, peak ball speed, 1.0\,m success rate, or 0.5\,m success rate.}
  \label{fig:results}
\end{figure}

\section{Additional Results}
\label{app:extra}

We plan the following diagnostics to further characterize the system beyond the main simulation and hardware evaluations:
\begin{itemize}\setlength\itemsep{1pt}
  \item \textbf{LiDAR ball-localization accuracy.} Position error vs.\ a motion-capture or tape-measure ground truth across the working range, and its effect on kick accuracy.
  \item \textbf{Proximity relaxation scale sensitivity.} Performance vs.\ a uniform scaling of all near-ball scales $\mu_i$ and vs.\ the $(d_\text{near},d_\text{far})$ window, to show robustness to these choices.
  \item \textbf{Whole-body coordination.} Joint-angle trajectories over a canonical kick cycle, showing the swing-leg hip/knee/ankle chain coupled with arm counter-rotation; comparison against the motion reference.
  \item \textbf{Extended hardware analysis.} Per-target failure-mode breakdown (perception failure / contact miss / balance loss) with Wilson confidence intervals, and additional kick-sequence montages including failure cases.
  \item \textbf{Weight-only ablation.} Stage-3 training with the ball-spawn range held fixed and only $w_g$ varied, to isolate the goal weight from the exploration-difficulty schedule.
\end{itemize}

\end{document}